\documentclass[journal]{IEEEtran}
\usepackage{cite}

\usepackage{graphicx}
\ifCLASSINFOpdf
\else
\fi
\usepackage{amsmath}
\usepackage{algorithm}
\usepackage{algorithmic}
\usepackage[caption=false,font=footnotesize]{subfig}

\usepackage{color}
\usepackage{booktabs}
\usepackage{url}

\begin{document}
%
\title{Visual Perception Generalization for Vision-and-Language Navigation via Meta-Learning}
%
%
%

\author{Ting Wang, 
        Zongkai Wu*, 
        and~Donglin Wang*
\thanks{T. Wang, Z. Wu and D. Wang are with the Machine Intelligence Lab (MiLAB) of the School of Engineering, Westlake University, Hangzhou, 310024 China e-mail: (wangting@westlike.edu.cn; wuzongkai@westlike.edu.cn; wangdonglin@westlike.edu.cn).}
\thanks{T. Wang is with the College of Computer Science and Technology, Zhejiang University, Hangzhou, 310024 China.}
\thanks{*Z. Wu and D. Wang are Co-corresponding authors.}}


%
%

\markboth{Journal of \LaTeX\ Class Files,~Vol.~14, No.~8, August~2015}%
{Shell \MakeLowercase{\textit{et al.}}: Bare Demo of IEEEtran.cls for IEEE Journals}
%



\maketitle

\begin{abstract}
Vision-and-language navigation (VLN) is a challenging task that requires an agent to navigate in real-world environments by understanding natural language instructions and visual information received in real-time.
Prior works have implemented VLN tasks on continuous environments or physical robots, all of which use a fixed camera configuration due to the limitations of datasets, such as 1.5 meters height, 90 degrees horizontal field of view (HFOV), etc.
However, real-life robots with different purposes have multiple camera configurations, and the huge gap in visual information makes it difficult to directly transfer the learned navigation skills between various robots. 
In this paper, we propose a visual perception generalization strategy based on meta-learning, which enables the agent to fast adapt to a new camera configuration with a few shots. 
In the training phase, we first locate the generalization problem to the visual perception module, and then compare two meta-learning algorithms for better generalization in seen and unseen environments. 
One of them uses the Model-Agnostic Meta-Learning (MAML) algorithm that requires a few shot adaptation, and the other refers to a metric-based meta-learning method with a feature-wise affine transformation layer. 
The experiment results on the VLN-CE dataset demonstrate that our strategy successfully adapts the learned navigation skills to new camera configurations, and the two algorithms show their advantages in seen and unseen environments respectively. 
\end{abstract}


%
\IEEEpeerreviewmaketitle

\section{Introduction}
%
%
%
%

\IEEEPARstart{T}{he} vision-and-language navigation (VLN) task requires an agent to follow natural language instructions to navigate in photo-realistic environments according to the visual information captured in real-time and the pre-built navigation graph \cite{anderson2018vision}. 
In recent years, the VLN task has attracted widespread attention due to its promising real-life applications and many methods have achieved satisfactory results in terms of success rate in simulation environments \cite{wang2019reinforced,tan2019learning,ma2019self}. 
Recently, \cite{krantz2020beyond} breaks through the limitation of the navigation graph and migrates the VLN task to continuous environments for the first time. 
Subsequently, combining a subgoal module with the traditional path planning on maps, \cite{anderson2020sim} transfers the VLN task from the embodied agent in simulation environments to the physical robot in real environments and successfully guarantees an acceptable success rate. 

\begin{figure}[t]
	\includegraphics[scale=0.57]{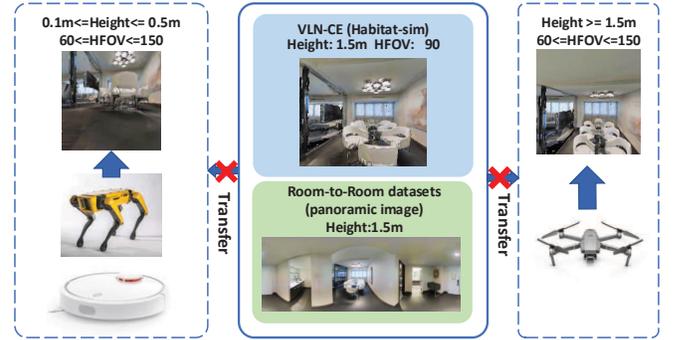}
	\centering
	\caption{Illustration of visual perception generalization problem. In the VLN task, the agent usually uses a fixed configuration camera, such as 1.5 meters high, panoramic view or 90 degree HFOV. However, in real life, robots with different functions and different forms apparently have very distinct camera configurations. The discrepancy in visual information obtained by different visual perceptions is very obvious, which makes it impossible to directly use the learned navigation skills on other robots. } 
	\label{fig1}
\end{figure}

However, due to the limitation of datasets, the above-mentioned agents or robots are trained by a fixed camera configuration, such as 1.5 meters height, 90 degrees horizontal field of view (HFOV) .etc.
But in fact, robots for different purposes have various forms, and their camera configurations are even more different. 
As shown in Figure \ref{fig1}, the camera heights or HFOVs of robots with specific functions are different such as sweeping robots, search and rescue drones, four-legged mechanical dogs, etc. 
The huge gap in visual information acquired by cameras with different configurations makes it difficult for learned navigation models to be directly shared among different robots. 

To solve the perception generalization problem of heterogeneous robots in VLN, we propose a visual perception generalization strategy based on meta-learning. 
According to the functions of each module of the VLN model, we divide the overall VLN model into a visual perception module, a language understanding module, and a navigation reasoning module.
Since camera configurations only affect the received image information, we locate the generalization problem to the visual perception module.
We hope that the visual perception module extracts the same features in the same location under different sensor configurations, so that the navigation module can make corresponding consistent action decisions. 
 
To train our visual perception module, we pre-train a VLN navigation model with a classic structure beforehand. 
Then, the learned visual perception module is used as the supervision information for our meta-learner. 
We propose two methods based on meta-learning to train our visual perception module, and compare the two methods in seen and unseen environments. 
For the first method, inspired by a few-shot adaptation of visual navigation \cite{luo2020few}, we train the visual perception module with the Model-Agnostic Meta-Learning (MAML) algorithm, so that the visual perception module can quickly adapt to a new sensor configuration through fine-tuning with a small amount of data.
However, there are two concerns about the MAML algorithm, one is that a small amount of data may still not be readily acquired, and the other is that the inaccuracy of the learned visual perception module in unseen environments may bring extra errors. 
Consequently, we consider another method, that is, adding feature-wise affine transformation (AT) layers to the visual perception module to simulate visual features under different sensor configurations. The AT parameters are trained by learning to learn, so that the visual perception module can be better generalized to arbitrary camera configuration without any adaptation process.

In the validation phase, for cameras with different heights and HFOVs, we evaluate the above two mentioned methods on the Habitat simulator involving vision-and-language navigation in continuous environments (VLN-CE) \cite{krantz2020beyond} tasks. 
The experiment results show that our strategy successfully adapts the learned navigation model to new sensor configurations, and the two methods show their advantages in seen and unseen environments respectively. 
It is worth mentioning that our methods are suitable for visual generalization between sensors of any different configurations. In this paper, we only explain from the aspect of camera height and HFOV. 

In general, our main contributions are as following:
\begin{itemize}
	\item We are the first to take notice of the visual generalization problem brought by different sensor configurations for different robots in VLN tasks, and locate the problem to the visual perception module. 
	
	\item We propose a visual perception generalization strategy based on meta-learning to deal with the generalization problem, where we compare the two methods for better generalization in seen and unseen environments respectively. 
	
	\item We implement the visual perception generalization under different camera heights and HFOVs. The experimental results prove the effectiveness of our strategy, and the two methods show their superiority in seen and unseen environments separately. 
\end{itemize}

\section{Related Work}
\subsection{Vision-and-Language Navigation}
\cite{anderson2018vision} first propose the concept of Vision-and-Language Navigation (VLN) and provide the Room-to-Room (R2R) dataset collected using the Matterport3D Simulator based on real images. On this basis, many works have made progress and addressed some of the challenges for the VLN task.  
\cite{fried2018speaker} propose a speaker model, which generates corresponding natural language instructions according to the sampled new paths for data augmentation and route selection. 
To tackle the problems of cross-modal grounding and ambiguous feedback, \cite{wang2019reinforced} propose a Reinforced Cross-Modal Matching (RCM) approach, using a matching critic as an intrinsic reward to encourage global matching between language instructions and routes, and using a reasoning navigator with three attention mechanisms to align the visual context, trajectories and language instructions for local scene navigation. 
\cite{wang2019reinforced} also introduce a Self-Supervised Imitation Learning (SIL) method, which imitates its own past good behaviors in an unseen environment for improving the generalization. 
\cite{ma2019self} propose a progress monitor to estimate the distance from the current view-point to the final target. 
\cite{wang2018look} first combine model-free and model-based deep reinforcement learning for VLN and show good performance in unseen environments. 
The main contribution of \cite{tan2019learning} is to create new environments by applying dropout on existing environments, and use the speaker model \cite{fried2018speaker} to generate natural language instructions corresponding to the sampled paths in the new environments to achieve data augmentation. 
\cite{zhu2020vision} introduce four self-supervised auxiliary reasoning tasks, which refer to the methods \cite{fried2018speaker,wang2019reinforced,ma2019self} proposed previously. 
\cite{zhu2020multimodal} propose a Multimodal Text Style Transfer (MTST) approach to deal with the data scarcity problem in outdoor navigation.
Many other methods such as \cite{zhang2020language,qi2020object,yu2020take,hao2020towards} also have been proposed to explore the VLN task from various angles and have made some progress. 

\cite{ku2020room} introduce a new dataset Room-Across-Room (RxR) for VLN. RxR is multilingual and larger than other VLN datasets, adding virtual gestures aligned in time with instructions, which provides new possibilities for the development of VLN. 
Eliminating the unrealistic perfect assumption based on the navigation graph, \cite{krantz2020beyond} first propose a new task, Vision-and-Language Navigation in Continuous Environments (VLN-CE) on top of the Habitat Simulator \cite{savva2019habitat}, and develop models as baselines. Our experiments are carried out under the setting of this article.
Afterward \cite{anderson2020sim} transfer the VLN task from the simulation environment to the physical robotic platform, which makes the application of VLN in real life a big step forward.


\subsection{Meta-Learning} 
Meta-learning hopes that the model can acquire a "learning to learn" ability, so that the model can quickly adapt to new tasks based on the knowledge learned from previous tasks. One remarkable meta-learning algorithm is Model-Agnostic Meta-Learning (MAML) \cite{finn2017model}, which enables the model to quickly and effectively adapt to new tasks using only a small amount of data by optimizing the initial values of parameters. To achieve this goal, the training model needs to maximize the parameter sensitivity of the loss function of the new task, hoping that very small changes of parameters can also bring great improvements to the model. 
MAML is suitable for many fields, including regression, image classification, and reinforcement learning, without being limited by the model itself. 

\cite{wortsman2019learning} and \cite{luo2020few} combine MAML and visual navigation. The former proposes a self-adaptive visual navigation (SAVN) method that enables the agent to interact with the environment without any additional supervision through meta-reinforcement learning. The latter divides the navigation framework into perception and inference networks, and utilizes MAML to train the perception network so that the agent can adapt to new observations with a few shots. 

In addition to the above optimization-based content, metric-based learning is also one of the common methods of meta-learning. A metric-based model usually consists of a feature encoder for feature extraction and a metric function for classification tasks. 
The Matching Net \cite{vinyals2016matching} uses Cosine in the embedding space to measure the features extracted from the support set, and achieves classification by calculating the matching degree on the test samples. Referring to the clustering idea, Prototypical Networks \cite{snell2017prototypical} projects the support set into a metric space to obtain the vector mean, and calculates the distance from the test sample to each prototype for classification. The relation module proposed by the Relation Network \cite{sung2018learning} replaces Cosine and Euclidean distance metric in the Matching Net and Prototypical Networks, making it a learnable nonlinear classifier for judging relations and realizing classification. These are three currently popular metric-based meta-learning models. 
%
We draw on the feature-wise affine transformation layers on the feature encoder to simulate the distribution of image features captured by cameras with different configurations to achieve the generalization of visual perception.

\section{Visual Perception Generalization for VLN}
In this section, we propose a visual perception generalization strategy, which can make the trained VLN agent generalize quickly and effectively to others agents with different sensor configurations. 

\begin{figure}[h]
	\includegraphics[scale=0.57]{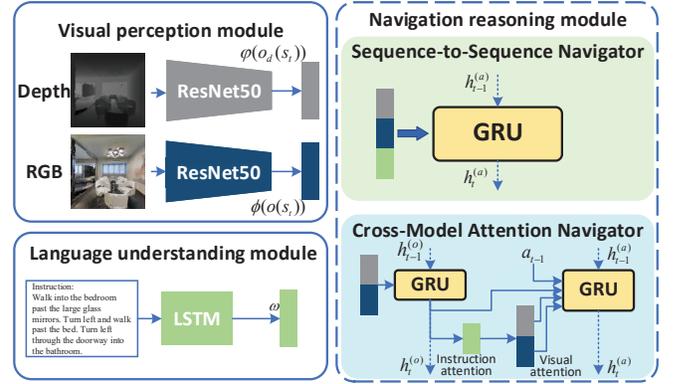}
	\centering
	\caption{Framework of the VLN model. We divide the complete VLN model into three modules: (1) a visual perception module; (2) a language understanding module; (3) a navigation reasoning module, where the upper right represents a simple sequence-to-sequence navigator, and the lower right represents a more complex cross-modal attention navigation. Our visual perception generalization strategy only modifies the visual perception module and fixes the other two parts. } 
	\label{fig2}
\end{figure}

As described by the framework of the VLN model in Figure \ref{fig2}, we divide the VLN model into a visual perception module, an instruction understanding module, and a navigation reasoning module. 
We locate the generalization problem to the visual perception module and assume that the action space is consistent for any agent. 
Hence, we only adapt the visual perception module to new sensor configurations based on meta-learning while freezing the instruction understanding module and the navigation reasoning module. 

\subsection{Task Formulation}

In the VLN task, given a natural language instruction, the agent's goal is to navigate towards the target location by co-processing language and image information inside real continuous indoor environments.

We consider each task as a triple $ \tau=(I, X, S) $ ,where $ I\in\mathcal{I} $ represents one of a series of environments, $ X=\{x_0,x_1,\dots,x_L\} (X \in \mathcal{X})$ is a natural language instruction with $ L $ words and $ S=\{s_0, s_1, \dots, s_t\} $ is a state space. 
Each agent has its own observation space, and can obtain RGB images $ o(s_t) $ and Depth images $ o_{d}(s_t) $ according to the current state $ s_t $ at each time $ t $ in the current environment $I$. $ o(\cdot) $ and $ o_{d}(\cdot) $ represent the specific observation functions related to the sensor configuration, which means that even at the same state, different camera configurations still provide agents with distinct image inputs and visual features. The diversity of observation functions prevents the navigation skills learned for the VLN agent from being generalized to a new observation space. 

Then the agent predicts the current action $ a \in \mathcal{A} $ based on the current state, vision (RGB and Depth), and instruction information. 
The action space $ \mathcal{A} $ of all agents consists of four simple, low-level actions - Move forward, Turn left or Turn right, and Stop, which is the default setting. The success of the VLN task can be declared only if the agent's position is close enough to the instruction target when it makes the "Stop" action.

\subsection{Visual Perception Generalization via meta-learning} 

To adapt our baseline agents to different observation spaces, the most straightforward way is to retrain a new navigation policy. However, retraining requires a lot of data, time, and labor costs. 
Therefore, we take a generalization strategy based on meta-learning to enable agents to adapt effectively to a new observation space. 

In our strategy, we have a learned VLN model with the training of a fixed observation space. 
Under this situation, we freeze the navigation reasoning module and language understanding module.
The learned visual perception module is used as the supervision information for retraining with meta-learning, which is to promote the visual perception module to produce as close as possible intermediate visual features under the same location but different observations, that is $ \tilde{\phi}(\tilde{o}(s_t)) \to \phi(o(s_t)) $ and $ \tilde{\varphi}(\tilde{o}_d(s_t)) \to \varphi(o_d(s_t)) $. 
$ \phi $ and $ \varphi $ represent the trained ResNet50, and $ \tilde{\phi} $ and $ \tilde{\varphi} $ represent the new ResNet50 we want to learn that can be used with new sensors of different configurations. 
Using the learned visual perception module as supervision, we can train a new visual perception module by minimizing the loss. 

\begin{equation}
\mathcal{L}_r = \Sigma |\phi(o(s_t))-\tilde{\phi}(\tilde{o}(s_t))|
\end{equation}

\begin{equation}
\mathcal{L}_d = \Sigma |\varphi(o_d(s_t))-\tilde{\varphi}(\tilde{o}_d(s_t))|
\end{equation}

Using supervised learning to optimize the visual perception module requires a large number of observations from the target observation space as training data, which makes pure supervised learning time-consuming and infeasible. 
Therefore, we use meta-learning to train the visual perception module, which requires just a few shots. 
Here we consider two meta-learning-based methods with pros and cons to train the visual perception module for generalization. 

\paragraph{Few-shot Adaptation with MAML}

We first propose an algorithm with Model-Agnostic Meta-Learning (MAML) \cite{finn2017model} framework, that can quickly adapt navigation agents to new sensors with a very small amount of data and a few shots fine-tuning. 

We consider the distribution of VLN tasks $ p(\mit \Gamma) $. Each support set of the task $ \mit \Gamma_i $ consists of $ k $ observation images with specific visual perception functions $ \tilde{\phi} $ or $ \tilde{\varphi} $.
We train our visual perception module with $ \tilde{\phi_\theta} $ and $ \tilde{\varphi_\mu} $, which is parameterized by $ \theta $ and $ \mu $ respectively, to learn an unseen task using only $ k $ samples. We randomly sample a batch of tasks $ \mit \Gamma_i (i = 1 \dots N)$ from the distribution and calculate adapted parameters $ \theta_i^\prime $, $ \mu_i^\prime $ with gradient descent. Then, the meta-parameters $ \theta $ and $ \mu $ are updated by the adapted parameters. The details of the algorithm are shown in Algorithm~\ref{alg:MAML}. 

\begin{algorithm}[t]
	\caption{Few-Shot Adaptation with MAML}
	\label{alg:MAML}
	\textbf{Require}: Task distribution $ p(\mit \Gamma) $\\
	\textbf{Require}: Learning rate $ \alpha $, $ \beta $, $ \gamma $, $ \delta $
	
	\begin{algorithmic}[1] 
		\STATE Randomly initialize $ \theta $ and $ \mu $
		\WHILE{not done}
		\FOR{mini-batch of tasks $ \mit \Gamma_i \in p(\mit \Gamma) $}
		\STATE Calculate $ \nabla_\theta \mathcal{L}_{r, \mit \Gamma_i} (\tilde{\phi}_\theta)$ and $ \nabla_\mu \mathcal{L}_{d, \mit \Gamma_i} (\tilde{\varphi}_\mu)$ 
		\STATE Compute adapted parameters with gradient descent: $ \theta_i^\prime = \theta-\alpha \nabla_\theta \mathcal{L}_{r, \mit \Gamma_i} (\tilde{\phi}_\theta) $ ,\\
		$ \mu_i^\prime = \mu-\beta \nabla_\mu \mathcal{L}_{d, \mit \Gamma_i} (\tilde{\varphi}_\mu) $
		\ENDFOR
		\STATE Update $ \theta = \theta-\gamma \nabla_\theta \Sigma_{\mit \Gamma_i \sim p(\mit \Gamma)}\mathcal{L}_{r, \mit \Gamma_i} (\tilde{\phi}_{\theta_i^\prime}) $ and \\
		\qquad \quad $ \mu = \mu-\delta \nabla_\mu \Sigma_{\mit \Gamma_i \sim p(\mit \Gamma)}\mathcal{L}_{d, \mit \Gamma_i} (\tilde{\varphi}_{\mu_i^\prime}) $
		\ENDWHILE
	\end{algorithmic}
\end{algorithm}

\paragraph{Generalization with Affine Transformation}

MAML-based optimization methods need to go through a few-shot adaptation process when facing new sensors, but obtaining even a little data requires manpower. To this end, referring to metric-based meta-learning, we propose another generalization method that does not require any adaptation means for VLN. 

We assume $ K $ domains of seen camera configurations $ \{\mathcal{T}_1^{seen}, \mathcal{T}_2^{seen}, \dots, \mathcal{T}_K^{seen}\} $ available in the training phase. Our goal is that the visual perception module learned from the existing senor configurations can generalize well to a new sensor configuration. For example, the model we trained using the visual information obtained by $ 0.5 $m, $ 1.0 $m, and $ 1.5 $m high cameras can still show good generalization performance at other camera heights. 

\begin{figure}[t]
	\includegraphics[scale=0.74]{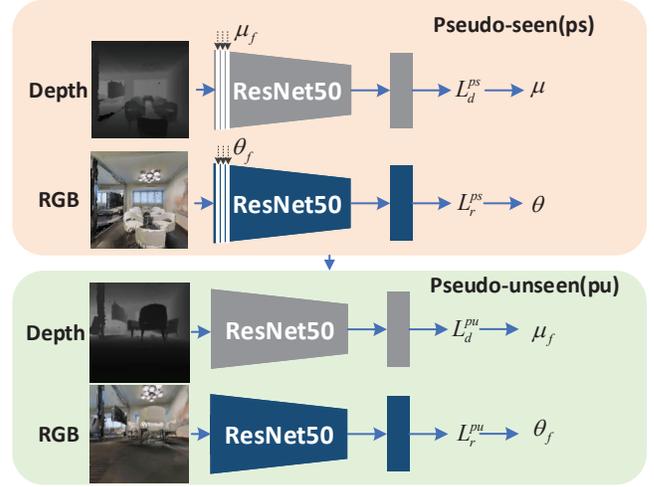}
	\centering
	\caption{Overview of the generalization with affine transformation layers. For the pseudo-seen task, we add the AT layers to the two ResNet50 networks to imitate the visual features of the image information obtained by agents with different configuration sensors and update the parameters $ \theta $, $ \mu $ of the ResNets. For the pseudo-unseen task, we remove the AT layers to measure the generalization performance of the visual perception module, and update the AT parameters $ \theta_f $, $ \mu_f $. } 
	\label{fig3}
\end{figure}

To address the above problem, we insert feature-wise affine transformation (AT) layers \cite{tseng2020cross} on the ResNets of the visual perception module to augment the image features for imitating various visual features from different high cameras, as illustrated in Figure \ref{fig3}.  
The hyper-parameters $ \{\theta_{\varepsilon},\theta_{\rho}\} \in R^{C \times 1 \times 1} $ and $ \{\mu_{\varepsilon},\mu_{\rho}\} \in R^{C \times 1 \times 1} $ indicate the standard deviations of the Gaussian distributions for sampling the AT parameters. Given an intermediate feature activation map $ z_\theta $ and $ z_\mu $in the ResNets with the dimension of $ C \times H \times W $, we first sample the scaling term $ \varepsilon $ and bias term $ \rho $ from Gaussian distributions,
\begin{equation}
\varepsilon_\theta \sim N(1,\text{softplus}(\theta_{\varepsilon})), \quad 
\rho_\theta \sim N(0,\text{softplus}(\theta_{\rho}))
\end{equation}
\begin{equation}
\varepsilon_\mu \sim N(1,\text{softplus}(\mu{\varepsilon})), \quad 
\rho_\mu \sim N(0,\text{softplus}(\mu{\rho})).
\end{equation}
We then compute the modulated activation $ \hat{z} $ as 
\begin{equation}
\hat{z}_\theta = \varepsilon_\theta \times z_\theta + \rho_\theta, \quad 
\hat{z}_\mu = \varepsilon_\mu \times z_\mu + \rho_\mu. 
\end{equation}
We use the way of learning to learn to optimize the hyperparameters $ \theta_f = \{ \theta_{\varepsilon},\theta_{\rho} \} $ and $ \mu_f = \{\mu_{\varepsilon},\mu_{\rho}\} $ of the AT layer, as described in Algorithm~\ref{alg:metric}. 
In each training iteration, we sample a pseudo-seen $ \mathcal{T}^{ps} $ and a psuedo-unseen $ \mathcal{T}^{pu} $ domain from a set of seen camera configurations $ \{\mathcal{T}_1^{seen}, \mathcal{T}_2^{seen}, \dots, \mathcal{T}_K^{seen}\} $. We then update the parameters in the ResNets with the pseudo-seen task $ \mit \Gamma^{ps} $, namely
\begin{equation}\label{11}
\theta = \theta - \eta \nabla_\theta \mathcal{L}_r^{ps} (\phi(o(s_t)), \tilde{\phi}_{\theta,\theta_f}(\tilde{o}(s_t)))
\end{equation}
\begin{equation}\label{12}
\mu = \mu - \zeta \nabla_\mu \mathcal{L}_d^{ps} (\varphi(o_d(s_t)), \tilde{\varphi}_{\mu,\mu_f}(\tilde{o}_d(s_t))). 
\end{equation}
where $ \eta $ and $ \zeta $ are learning rates. We measure the generalization performance of the updated visual perception module by (a) removing the AT layers from the ResNets and (b) computing the loss $ \mathcal{L}_r^{pu} $ and $ \mathcal{L}_d^{pu} $ of visual difference on the pseudo-unseen task $ \mit \Gamma^{pu} $, which reflect the effectiveness of the AT layers. Finally, we update $ \theta_f $ and $ \mu_f $ by  
\begin{equation}\label{13}
\theta_f = \theta_f - \eta \nabla_{\theta_f} \mathcal{L}_r^{pu}(\phi(o(s_t)), \tilde{\phi}_{\theta}(\tilde{o}(s_t)))
\end{equation}
\begin{equation}\label{14}
\mu_f = \mu_f - \zeta \nabla_{\mu_f} \mathcal{L}_d^{pu}(\varphi(o_d(s_t)), \tilde{\varphi}_{\mu}(\tilde{o}_d(s_t))). 
\end{equation}

\begin{algorithm}[t]
	\caption{Learning-to-Learn Affine Transformation}
	\label{alg:metric}
	\textbf{Require}: Seen perspectives $ \{\mathcal{T}_1^{seen}, \mathcal{T}_2^{seen}, \dots, \mathcal{T}_K^{seen}\} $\\
	\textbf{Require}: Learning rate $ \eta $ and $ \zeta $
	\begin{algorithmic}[1] 
		\STATE Randomly initialize $ \theta $, $ \mu $, $ \theta_f $ and $ \mu_f $
		\WHILE{training}
		\STATE Randomly sample non-overlapping pseudo-seen $ \mathcal{T}^{ps} $ and psuedo-unseen $ \mathcal{T}^{pu} $ domains from $ p(\mathcal{T}_K^{seen}) $
		\STATE Sample a pesudo-seen task $ \mit \Gamma^{ps} \in \mathcal{T}^{ps} $ and a pseudo-unseen task $ \mit \Gamma^{pu} \in \mathcal{T}^{pu} $
		\STATE Update the parameters $ \theta $ and $ \mu $ of the visual perception module with the pseudo-seen task using equation~\ref{11},~\ref{12}
		\STATE Update the parameters $ \theta_f $ and $ \mu_f $ of the feature-wise affine transformation layers with the pseudo-unseen task using equation~\ref{13},~\ref{14}
		\ENDWHILE
	\end{algorithmic}
\end{algorithm}

\subsection{Language Understanding and Navigation reasoning baseline}
In our task, the language understanding and navigation reasoning module are not affected by the change of observation space.
Hence, we select two models mentioned in the paper \cite{krantz2020beyond} for VLN in continuous environments as our baselines, that can understand the natural language instructions and navigate to the target location smoothly based on language and vision input. 

\paragraph{Sequence-to-Sequence Model} 
As shown in the upper part of the navigation reasoning module in Figure \ref{fig2}, the core of the simple sequence-to-sequence model is a Gate Recurrent Unit (GRU) \cite{cho2014learning}, which takes visual information and language instructions as input and predicts the next action. 
Through the visual perception module, the semantic visual features $ \phi(o(s_t)) $ (for RGB) and depth information $ \varphi(o_{d}(s_t)) $ (for Depth) can be obtained respectively, and the last hidden state of LSTM \cite{hochreiter1997long} is applied to encode language instructions as $ \omega = \text{LSTM} (x_1,x_2,\dots,x_L) $. At time step t, the predicted action $ a_t $ of the VLN agent is expressed as 
\begin{equation}
h_t^{(a)} = \text{GRU} ([\phi(o(s_t)), \varphi(o_{d}(s_t)), \omega], h_{t-1}^{(a)})
\end{equation}

\begin{equation}
a_t = \mathop{\arg\max}_{a} \ \text{softmax}(W_a h_t^{(a)}+b_a)
\end{equation}

\paragraph{Cross-Model Attention} 
Based on the above model, a more complex model with two recurrent neural networks is shown in the lower right part of Figure \ref{fig2}. One GRU is used to track previous visual observations, and the other is used to predict actions based on attentioned language and visual information. For the first GRU,
\begin{equation}
h_t^{(o)} = \text{GRU} ([\phi(o(s_t)),\varphi(o_{d}(s_t)),a_{t-1}],h_{t-1}^{(o)})
\end{equation}
where $ a_{t-1} $ is a linear embedding of the previous action. Here we use a bidirectional LSTM \cite{schuster1997bidirectional} to encode language instructions and express them as $ \Omega = \{\omega_1,\omega_2,\dots,\omega_L\} = \text{Bi-LSTM}(x_1,x_2,\dots,x_L) $. 
Then the latter GRU outputs an action $ a_t $ as
\begin{equation}
\resizebox{.91\linewidth}{!}{$
	\displaystyle
	h_t^{(a)} = \text{GRU} ([\hat{\omega_t}, \hat{\phi(o(s_t))}, \hat{\varphi(o_{d}(s_t))}, a_{t-1}, h_t^{(o)}], h_{t-1}^{(a)})
	$}
\end{equation}
\begin{equation}
a_t = \mathop{\arg\max}_{a} \ \text{softmax}(W_a h_t^{(a)}+b_a)
\end{equation}
where $ \hat{\omega_t} = \text{Att}(\Omega,h_t^{(o)}) $, $ \hat{\phi(o(s_t))} = \text{Att}(\phi(o(s_t)),\hat{\omega_t}) $ and $ \hat{\varphi(o_{d}(s_t))} = \text{Att}(\varphi(o_{d}(s_t)), \hat{\omega_t}) $ that are the results after calculating the attention. 

\section{Experiments}

\subsection{Dataset}
We use the VLN-CE dataset \cite{krantz2020beyond} to evaluate our visual perception generalization strategy, which is for the VLN task rebuilt upon continuous Matterport3D environments \cite{chang2017matterport3d} in the Habitat simulator \cite{savva2019habitat}. The VLN-CE dataset collects 4475 trajectories converted from the R2R dataset \cite{anderson2018vision} in 90 buildings, each of which is described by three natural language instructions with an average of 29 words and provided a pre-computed shortest path via low-level actions (Move forward $ 0.25m $, Turn left or Turn right $ 15 $ degrees, and Stop). Due to the low-level action space, the average length of the trajectories is $ 55.88 $ steps.
The VLN-CE dataset is divided into the training set, the validation set and the test set. 
The validation set is further split into two parts: the validation-seen subset, where the paths are sampled from scenes appeared in training set, and the validation-unseen subset, where samples are from unseen environments. 

\subsection{Metrics}

We evaluate the performance for our visual perception generalization strategy using five metrics: the trajectory length (\textbf{TL}), the navigation error (\textbf{NE}), the oracle rate (\textbf{OR}), the success rate (\textbf{SR}) and the success rate weighted by the path length (\textbf{SPL}). Specific values refer to \cite{krantz2020beyond}.
\begin{itemize}
	\item  \textbf{TL} measures the average length of the predicted trajectories in navigation. 
	\item \textbf{NE} measures the average distance (in meter) between the agent's stopping position in the predicted trajectory and the goal in the reference trajectory. 
	\item \textbf{OR} is the proportion of the closest point in the predicted trajectory to the target in the reference trajectory within a threshold distance.
	\item \textbf{SR} is the proportion of the agent stopping in the predicted route within a threshold distance of the goal in the reference route. 
	\item \textbf{SPL} is a comprehensive metric method integrating SR and TL that takes both effectiveness and efficiency into account. 
\end{itemize}

\subsection{Implementation Setup}
We implement our agent on the Habitat simulator \cite{krantz2020beyond}. 
We download the models with the best results in the open-access website as our baseline parameters. 
For sequence-to-sequence baseline, DAgger-based \cite{ross2011reduction} training has the best result, for which the $ n $th set is collected by taking the oracle action with probability $ 0.75^n $ and the current policy action otherwise. 
For cross-model attention model baseline, the model with the progress monitor\cite{ma2019self}, DAgger ( with probability $ 0.75^{n+1} $) and data augmentation performs best. 

\paragraph{Experiments for MAML} 
For adaptation with MAML, our method implementation uses the same inner learning rate of $ 2e-4 $ and outer learning rate of $ 2e-4 $, three-shots, ten gradient step updates, and the Adam optimizer. 
We retrain the visual perception module using agents with $ 90 $ degree HFOV and three different height (ie $ 0.5 $m, $ 1.0 $m, $ 1.5 $m) cameras, or $ 1.5 $m height and three HFOVs (ie $ 90 $ degree, $ 120 $ degree, $ 150 $ degree) cameras. 
The three RGB and depth images obtained by the three agents with the same camera walking one step according to a language instruction in three random environments are used as the support set of a task $ \mit \Gamma_i $. We verify our method with adaptation on an agent with $ 0.2 $m, $ 90 $ degree HFOV or $ 1.5 $m, $ 60 $ degree HFOV camera in seen and unseen validation environments respectively. 

\begin{figure*}[t]
	\includegraphics[scale=0.57]{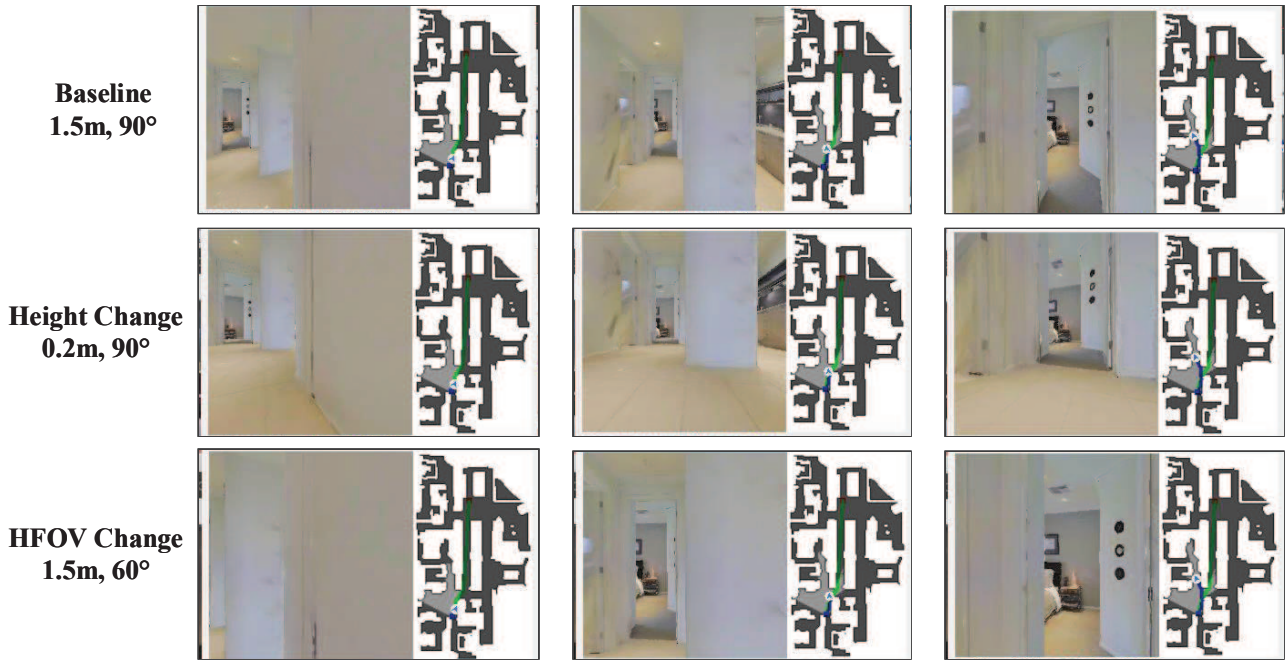}
	\centering
	\caption{Display of visual perception differences. We show the observation images of three cameras with different configurations in the same position. } 
	\label{fig4}
\end{figure*}

\paragraph{Experiments for Affine Transformation} 
For generalization with feature-wise affine transformation (AT) layers, our method implementation also uses the learning rate of $2e-4$ and the Adam optimizer. 
All seen visual perceptions $ \mathcal{T}_D^{seen} (D=\{1, 2, 3\})$ consist of observations obtained by agents with $ 0.5 $m, $ 1.0 $m, $ 1.5 $m height and $ 90 $ degree HFOV cameras, or $ 90 $ degree, $ 120 $ degree, $ 150 $ degree HFOV and $ 1.5 $m height cameras. Then, we randomly sample the pseudo-seen $ \mathcal{T}^{ps} $ and psuedo-unseen $ \mathcal{T}^{pu} $ from $ \mathcal{T}_D^{seen} $ in each training episode. 
We test our method without any adaptive process in seen and unseen validation environments on an agent with $ 0.2 $m, $ 90 $ degree HFOV or $ 1.5 $m, $ 60 $ degree HFOV camera respectively. The different camera configurations of agents during the training and testing phase are clearly stated in the table \ref{Table1}. The visual perception differences of cameras with different configurations are shown in the figure \ref{fig4}.

\begin{table}[t]
	\centering
	\caption{Experimental setup of the agent camera.}
	\label{Table1}
	\begin{tabular}{ccccccccccc}
		\toprule 
		\quad &training&testing\\
		\quad &height\quad HFOV&height\quad HFOV \\
		\midrule
		Generalization&1.5m\quad 90 degree&\quad\\
		for&1.0m\quad 90 degree&0.2m\quad 90 degree\\
		\textbf{camera height}&0.5m\quad 90 degree&\quad\\
		\midrule
		Generalization&1.5m\quad 90 degree&\quad\\
		for&1.5m\quad 120 degree&1.5m\quad 60 degree\\
		\textbf{camera HFOV}&1.5m\quad 150 degree&\quad\\
		\bottomrule 
	\end{tabular} 
\end{table}

\subsection{Comparison of Our Generalization Methods}
The validation dataset that is divided into two aspects based on seen and unseen environments is used to verify the effectiveness of our strategy. For validation-seen environments, the validation subset shares the same scenes with the training set but has new instructions. For validation-unseen environments, both instructions and scenes have no overlapping with the training dataset. 

In order to confirm the existence of generalization problem on different camera configurations, we first test the baselines with the validation dataset. 
Table \ref{Table2} summaries the results of our baselines in validation unseen environments. 
We train the sequence-to-sequence (seq2seq) and cross-model attention (CMA) baselines on an agent with a $ 1.5 $m height, $ 90 $ degree HFOV camera. 
Then we test our baselines for agents with $ 0.2 $m height, $ 90 $ degree HFOV camera (the upper part of the table \ref{Table2}) and $ 1.5 $m height, $ 60 $ degree HFOV camera (the bottom half of the table \ref{Table2}), respectively. 
Obviously, whether it is changing the agent's camera height or HFOV, all of the navigation performance is significantly reduced. 
This shows that there are huge gaps in visual perception under different camera configurations, which makes the navigation policies hard to effectively generalize to an agent or a robot with different camera configurations. 

Table \ref{Table3} illustrates the experiment results of using our generalization strategy compared with baselines in the validation seen and unseen environments, where five metrics are used for evaluation. Compared with the baselines, our strategy both achieve a competitive improvement regardless of changing the camera height or HFOV. 

For changing the camera height from $ 1.5 $m to $ 0.2 $m in validation-unseen environments, the MAML-based adaptation method achieves the success rates of $ 0.14 $ and $ 0.23 $, SPL of $ 0.12 $ and $ 0.20 $ on the two baselines, seq2seq and CMA, respectively. And the method based on AT layers respectively reach SR of $ 0.14 $ and $ 0.24 $, SPL of $ 0.12 $ and $ 0.20 $. The success rates of the two methods we introduce are much higher than the baselines. 
Although the results of the two methods are very similar to each other, it can be seen that the method with AT layers performs slightly better than the MAML-based method in unseen environments, while the MAML-based adaptation method performs better in seen environments. 
We speculate that the difference of performance in different environments may be caused by the few-shot adaptation of MAML. During the adaptation process, the gaps between seen and unseen environments cause the bias of results. 
The method that adds AT always shows relatively stable generalization performance because it simulates image features obtained by cameras with different unseen configurations and does not require any adaptation. 
For changing the camera HFOV from $ 90 $ degree to $ 60 $ degree, the performance of our generalization strategy is similar to the above analysis.

\begin{table}[!t]
	\centering
	\caption{Experimental results of VLN Baselines. The arrow direction indicates the optimization trend.}
	\label{Table2}
	\begin{tabular}{lcccccccccc}
		\toprule 
		\textbf{Change Height}&Validation-unseen\\
		($\text{1.5m}\rightarrow\text{0.2m}$)&TL$\downarrow$\quad\ NE$\downarrow$\quad\ OR$\uparrow$\quad\ SR$\uparrow$\quad\ SPL$\uparrow$ \ \ \\
		\midrule
		Seq2seq (1.5m)&8.46\quad 7.92\quad 0.35\quad 0.26\quad 0.23\\
		Seq2seq (0.2m)&6.99\quad 9.10\quad 0.11\quad 0.02\quad 0.01\\
		\midrule
		CMA (1.5m)&8.64\quad 7.37\quad 0.40\quad 0.32\quad 0.30\\
		CMA (0.2m)&9.44\quad 9.59\quad 0.12\quad 0.04\quad 0.03\\
		\bottomrule 
	\end{tabular} 
	\begin{tabular}{lcccccccccc}
		\toprule 
		\textbf{Change HFOV}&Validation-unseen\\
		($\text{90}^\circ\rightarrow\text{60}^\circ$)&TL$\downarrow$\quad\ NE$\downarrow$\quad\ OR$\uparrow$\quad\ SR$\uparrow$\quad\ SPL$\uparrow$ \ \ \\
		\midrule
		Seq2seq ($\text{90}^\circ$)&8.46\quad 7.92\quad 0.35\quad 0.26\quad 0.23\\
		Seq2seq ($\text{60}^\circ$)&9.21\quad 8.82\quad 0.20\quad 0.10\quad 0.08\\
		\midrule
		CMA ($\text{90}^\circ$)&8.64\quad 7.37\quad 0.40\quad 0.32\quad 0.30\\
		CMA ($\text{60}^\circ$)&8.96\quad 9.27\quad 0.23\quad 0.14\quad 0.12\\
		
		\bottomrule 
	\end{tabular} 
\end{table}

\begin{table*}[th]
	\centering
	\caption{Experiment results of our visual perception generalization strategy.}
	\label{Table3}
	\begin{tabular}{lcccccccccc}
		\toprule 
		&Validation-seen&Validation-unseen\\
		\textbf{Change Height:0.2m}&TL$\downarrow$\quad NE$\downarrow$\quad OR$\uparrow$\quad SR$\uparrow$\quad SPL$\uparrow$&TL$\downarrow$\quad NE$\downarrow$\quad OR$\uparrow$\quad SR$\uparrow$\quad SPL$\uparrow$\\
		\midrule
		Seq2seq baseline&\textcolor{black}{8.57\quad 10.1\quad 0.13\quad 0.04\quad 0.01}&6.99\quad 9.10\quad 0.11\quad 0.02\quad 0.01\\
		Seq2seq+MAML&\textcolor{black}{11.2\quad 9.13\quad 0.32\quad \textbf{0.13}\quad 0.11}&10.6\quad 8.89\quad 0.30\quad \textbf{0.14}\quad 0.12\\
		Seq2seq+AT&\textcolor{black}{9.52\quad 9.41\quad 0.25\quad 0.12\quad 0.11}&\textcolor{black}{9.59\quad 8.88\quad 0.26\quad \textbf{0.14}\quad 0.12}\\
		\midrule
		CMA baseline&\textcolor{black}{8.49\quad 8.98\quad 0.23\quad 0.08\quad 0.04}&9.44\quad 9.59\quad 0.12\quad 0.04\quad 0.03\\
		CMA+MAML&9.70\quad 8.49\quad 0.39\quad \textbf{0.28}\quad 0.25&9.02\quad 8.16\quad 0.34\quad 0.23\quad 0.20\\
		CMA+AT&\textcolor{black}{9.37\quad 8.49\quad 0.37\quad 0.26\quad 0.23}&\textcolor{black}{8.93\quad 8.15\quad 0.34\quad \textbf{0.24}\quad 0.22}\\
		
		\bottomrule 
	\end{tabular} 
	\begin{tabular}{lcccccccccc}
		\toprule 
		&Validation-seen&Validation-unseen\\
		\textbf{Change HFOV:$\text{60}^\circ$}&TL$\downarrow$\quad NE$\downarrow$\quad OR$\uparrow$\quad SR$\uparrow$\quad SPL$\uparrow$&TL$\downarrow$\quad NE$\downarrow$\quad OR$\uparrow$\quad SR$\uparrow$\quad SPL$\uparrow$\\
		\midrule
		Seq2seq baseline&\textcolor{black}{8.72\quad 8.59\quad 0.17\quad 0.13\quad 0.12}&9.21\quad 8.82\quad 0.20\quad 0.10\quad 0.08\\
		Seq2seq+MAML&\textcolor{black}{9.84\quad 9.17\quad 0.28\quad \textbf{0.21}\quad 0.19}&\textcolor{black}{10.2\quad 9.03\quad 0.25\quad 0.17\quad 0.15}\\
		Seq2seq+AT&\textcolor{black}{9.61\quad 9.03\quad 0.26\quad 0.19\quad 0.17}&\textcolor{black}{9.15\quad 8.59\quad 0.25\quad \textbf{0.18}\quad 0.16}\\
		\midrule
		CMA baseline&\textcolor{black}{8.72\quad 8.59\quad 0.22\quad 0.18\quad 0.10}&8.96\quad 9.27\quad 0.23\quad 0.14\quad 0.12\\
		CMA+MAML&\textcolor{black}{9.05\quad 8.34\quad 0.38\quad \textbf{0.28}\quad 0.25}&\textcolor{black}{9.76\quad 9.35\quad 0.33\quad 0.25\quad 0.22}\\
		CMA+AT&\textcolor{black}{8.37\quad 8.16\quad 0.32\quad 0.26\quad 0.24}&\textcolor{black}{9.01\quad 8.32\quad 0.36\quad \textbf{0.26}\quad 0.23}\\
		
		\bottomrule 
	\end{tabular} 
\end{table*}

\section{Conclusion}
This is the first work that pays attention to the generalization problem for the different sensor configurations in VLN tasks.
To solve this problem, we introduce a visual perception generalization strategy based on meta-learning that divides the VLN model into three modules (visual perception module, language understanding module, and navigation reasoning module), and locate the problem to the visual perception module. 
Then, we consider the MAML algorithm for an agent to adapt to a new camera configuration with a few-shot finetuning.
Due to the drawback of MAML in unseen environments, we propose another method using learning-to-learn that trains the visual perception module with adding affine transformation layers and tests without any adaptation.
Experimental results show that our generalization strategy has successfully generalized the learned visual perception module to a new camera configuration.
Besides, the two considered methods show their advantages in seen and unseen environments respectively. 
The MAML algorithm works better when there are a few shots for adaptation, but the algorithm with the affine transformation layer works slightly better when there is no data at all.


%

%

%

\ifCLASSOPTIONcaptionsoff
  \newpage
\fi



\bibliographystyle{IEEEtran}
\bibliography{IEEEexample}
\end{document}